\documentclass[conference]{IEEEtran}
\IEEEoverridecommandlockouts

\usepackage{cite}
\usepackage[table]{xcolor}

\definecolor{groupGray}{HTML}{E4E7EB}
\definecolor{groupBlue}{HTML}{DCEAF7}
\definecolor{groupGreen}{HTML}{DDEEDC}
\usepackage{amsmath,amssymb,amsfonts}
\usepackage{graphicx}
\usepackage{textcomp}
\usepackage{xcolor}
\usepackage{booktabs}
\usepackage{multirow}
\usepackage{makecell}
\usepackage{array}
\usepackage{url}
\newcommand{\skill}[1]{\texttt{#1}}
\newcommand{\up}{\,$\uparrow$}
\newcommand{\down}{\,$\downarrow$}

\begin{document}

\title{SurgiCLAW: An Interpretable Atomic-Skill Agent for Precision Robotic Micromanipulation}

\author{\IEEEauthorblockN{Angen Ye\textsuperscript{\dag}, Junjie Xie\textsuperscript{\dag},
Yujia Song, Junkai Huang, Dapeng Zhang\textsuperscript{*}}
\IEEEauthorblockA{Institute of Automation, Chinese Academy of Sciences\\
University of Chinese Academy of Sciences\\
Beijing, China\\
\textsuperscript{\dag}These authors contributed equally to this work.\\
\textsuperscript{*}Corresponding author.}
\IEEEauthorblockA{}}

\maketitle

\begin{abstract}
Recent multimodal large language models (MLLMs) and vision--language--action models have provided robots with new capabilities for long-horizon task understanding and planning. In surgical robotics, however, end-to-end prediction of low-level actions remains limited by insufficient precision, opaque decisions, and scarce data. This paper presents SurgiCLAW, a hierarchical skill-based framework for precision surgical manipulation. SurgiCLAW confines an MLLM to intraoperative visual understanding, structured explanation, and atomic surgical-skill selection, while delegating physical execution to verifiable deterministic control programs. For brain--machine-interface implantation, a representative high-precision surgical task, we construct a state-to-skill invocation dataset and expand it through visual and state perturbations. The model first learns structured skill invocation through supervised fine-tuning and is then optimized using three rewards for action correctness, operation prerequisites, and format validity, improving action selection, operational safety, and output parseability. We evaluate subtask success, offline model comparison, real-robot performance, and ablations. On a fixed test set, SurgiCLAW achieves 94.6\% skill-selection accuracy and 93.0\% macro-averaged F1 (Macro-F1). Across 50 closed-loop real-robot implantation trials, it achieves an 88.0\% task success rate.
\end{abstract}


\section{Introduction}
The rapid development of multimodal large language models and vision--language--action (VLA) models is changing the design paradigm of robotic systems\cite{kawaharazuka2025vision,intelligence2025pi_}. Compared with conventional end-to-end imitation-learning or reinforcement-learning policies, VLA models can extract high-level semantics from natural-language instructions, visual observations, and task context, showing promise in long-horizon planning, open-vocabulary interaction, and cross-scene generalization. This progress has motivated growing interest in using large models in surgical robots to understand intraoperative scenes, generate procedural plans, and assist with complex surgical tasks\cite{xu2025surgical}.

Surgical robotics, however, is fundamentally different from manipulation in ordinary open environments. The challenge is not only to decide what to do next, but also to execute the decision with sufficient precision, safety, and interpretability. Small needle-tip deviations, instrument-pose errors, or abnormal motion paths may cause tissue damage, bleeding, or other medical risks. Directly transferring the general VLA paradigm to surgical robots therefore has three limitations. First, end-to-end VLA models commonly predict continuous actions or low-level commands, and their local precision is difficult to reconcile with surgical error tolerances. Second, large-model outputs are often opaque: clinicians cannot easily determine why an action was selected, which visual and robot states support it, or whether current safety constraints are satisfied. Third, surgical data are inherently scarce. Acquiring precise manipulation trajectories requires domain experts, surgical environments, and high-accuracy synchronized recording, making large-scale real interaction data difficult to obtain.
\begin{table*}[t]
\caption{Atomic Action Interface}
\label{tab:skills}
\centering
\small
\begin{tabular}{p{0.23\textwidth}p{0.25\textwidth}p{0.43\textwidth}}
\toprule
\textbf{Function} & \textbf{Meaning} & \textbf{Primary prerequisite} \\
\midrule
\skill{ADJUST\_EXPOSURE} & Adjust microscopic exposure & Camera online; exposure outside the calibrated range \\
\skill{ADJUST\_FOCUS} & Adjust microscopic focus & Camera online; image out of focus \\
\skill{E\_STOP} & Trigger a safety stop & Singular configuration, system fault, or other high-risk anomaly \\
\skill{SPATIAL\_REGISTRATION} & Register image and robot coordinates & Usable image; system self-check passed \\
\skill{RESTART\_CAMERA} & Restart a failed or disconnected camera & Abnormal camera connection; robot held in a safe state \\
\skill{ALIGN\_NEEDLE} & Deterministic visual-servo needle alignment & Registration complete; image sharp; no singular configuration \\
\skill{INSERT} & Execute flexible-electrode insertion & Registration and alignment complete; camera online; all safety checks passed \\
\bottomrule
\end{tabular}
\vspace{-0.5em}
\end{table*}
We propose SurgiCLAW, an interpretable atomic-skill agent for precision surgical robotics. Its central principle is to decouple complex semantic decisions from high-precision control. An MLLM interprets images, robot states, and task instructions, produces a reviewable structured rationale, and selects an atomic surgical program. Execution is performed by predefined atomic-skill controllers. Each skill has explicit inputs, prerequisites, motion boundaries, and safety constraints, and can use geometry, optimization, visual servoing, or trajectory planning for precise execution. SurgiCLAW consequently does not ask a large model to control an instrument directly; it trains the model to perform an auditable state-to-skill mapping while verifiable controllers retain responsibility for precision motion.

This design also changes how training data are constructed. Instead of collecting complete continuous surgical trajectories, the task is formulated as supervised mapping from an image, robot state, and query to an atomic-skill invocation. Determining the appropriate atomic action and its essential parameters is generally easier, more consistent, and easier to audit than demonstrating a complete high-precision trajectory. For brain--machine-interface implantation, we propose a multi-MLLM collaborative data-generation method that expands 5,000 manually collected state--skill samples to approximately 30,000 structured training samples. A structured intermediate decision basis is also generated for every sample, so the output contains not only the function call but also the state assessment, visual evidence, task stage, and safety check.

The contributions are threefold:
\begin{enumerate}
    \item We introduce SurgiCLAW, a hierarchical atomic-skill agent for high-precision surgery. It confines the MLLM to intraoperative visual understanding, structured decision explanation, and atomic-skill selection, and connects it through a safety parser to deterministic controllers, separating high-level semantic decisions from low-level precision execution.
    \item We propose a multi-MLLM collaborative data-generation method for learning atomic surgical skills. Visual and state perturbations, candidate generation by multiple MLLMs, rule-based pre-validation, and cross-model verification allow high-quality state--skill training samples to be generated at scale from a limited set of robot images and state records.
    \item We design three rewards---action correctness, operation prerequisites, and format validity---and optimize the policy with group reward-decoupled normalization policy optimization (GDPO)\cite{liu2026gdpo}, systematically evaluating the method offline, on atomic subtasks, in a real-robot closed loop, and through ablation. SurgiCLAW achieves 94.6\% accuracy and 93.0\% Macro-F1 on the fixed test set and an 88.0\% task success rate in real-robot trials.
\end{enumerate}

\section{Related Work}
\subsection{Vision-Language-Action Policies for Robot Control}
Vision-language-action models couple web-scale semantic understanding with robot control: SayCan grounds language plans in executable skill affordances \cite{ahn2022can}, while RT-2 \cite{brohan2023rt} and OpenVLA \cite{kim2024openvla} predict actions directly from a unified vision-language model. These policies generalize across instructions and scenes, but emit continuous or token-level actions and rely on large amounts of interaction data. In a safety-critical, high-precision setting this raises two concerns that the present work targets: the predicted low-level motion cannot be guaranteed to meet a tight error bound, and the decision is opaque, so a clinician cannot inspect why an action was chosen. SurgiCLAW therefore keeps the semantic reasoning of a multimodal model but constrains its output to a small set of bounded, verifiable skills, so the model decides \emph{what} to do while verified controllers guarantee \emph{how precisely} it is done.

\subsection{Scene Understanding in Medical Robotics}
Autonomous surgical robotics spans supervised autonomy (STAR \cite{shademan2016supervised}), imitation-based control (SRT \cite{kim2024surgical}, SRT-H \cite{kim2025srt}), and language-guided assistance (SuFIA \cite{moghani2024sufia}); a parallel line builds medical vision-language models such as SurgVLM \cite{zeng2025surgvlm} and SurgVidLM \cite{wang2025surgvidlm} for operative scene understanding and query answering. These lines advance either task \emph{autonomy} or scene \emph{semantic interpretation}, but neither provides a safety-checked interface that converts a model's high-level decision into a precise, auditable motion primitive. Geometric-vision controllers for registration, alignment, and insertion \cite{ye2024non,ye2026fully,ye2026holenet} already supply such verified primitives for implantation; SurgiCLAW adds an adaptive, safety-gated decision layer that lets a large model schedule these primitives without directly controlling the end-effector.

\begin{figure*}[t]
\centering
\includegraphics[width=\textwidth]{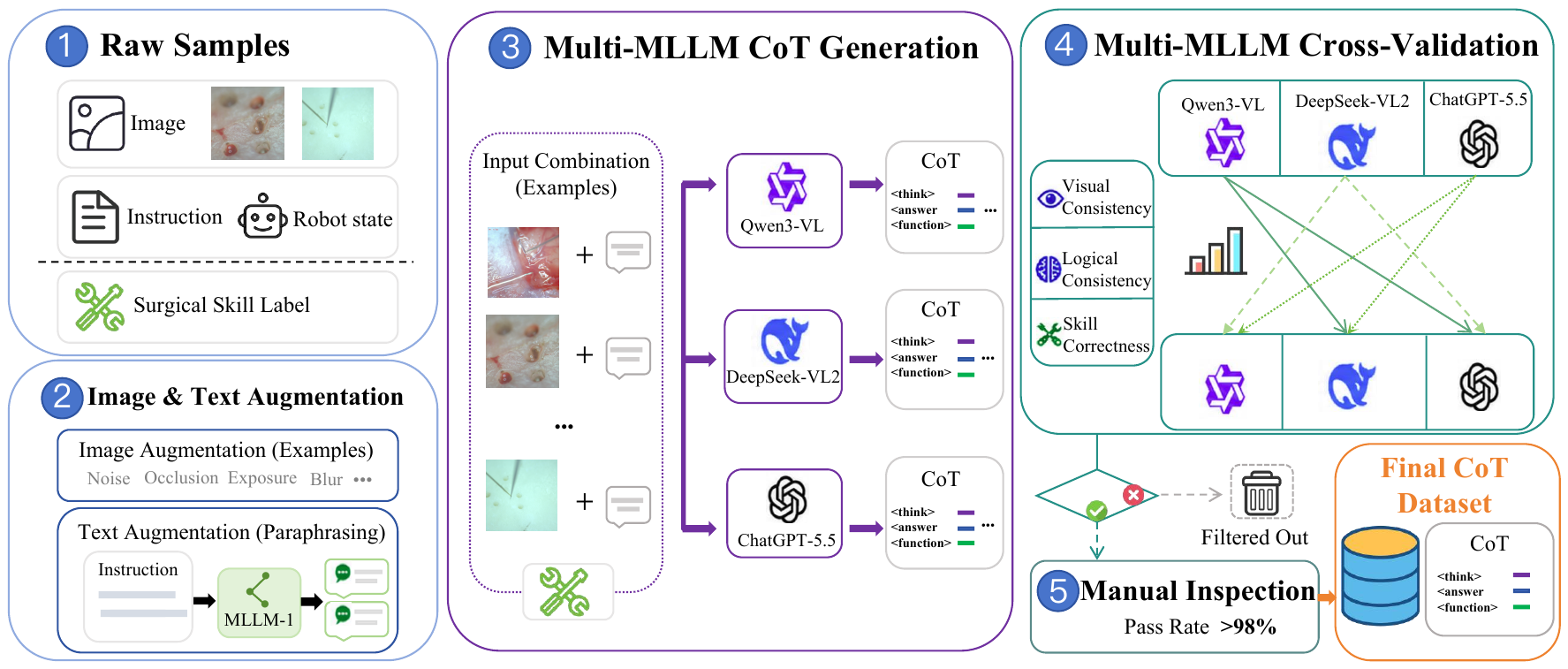}
\caption{Multi-MLLM collaborative data-generation pipeline. Starting from raw image, robot-state, instruction, and skill-label samples, the pipeline expands the data through image/text augmentation and constructs structured rationales using three MLLMs. Candidate outputs are filtered by rule-based checks, leave-one-generator-out cross-validation, and human spot inspection before being retained in the final structured dataset.}
\label{fig:data}
\vspace{-0.5em}
\end{figure*}

\section{Method}
\subsection{Hierarchical Architecture and Action Interface}

At decision step $t$, the system input is
\begin{equation}
x_t=\{I_t,s_t,q_t\},
\label{eq:input}
\end{equation}
where $I_t$ is the microscopic image, $s_t$ is the robot state, and $q_t$ is the current operation query. The state $s_t$ encodes the current joint values together with the non-visual safety conditions measured by the system---spatial-registration completion, camera connectivity, and robot singularity. The visual conditions that determine the current sub-skill, such as image-quality anomalies (blur, exposure), needle alignment, and scene usability, are perceived by the model directly from $I_t$. Each step is inferred independently without relying on an implicit dialogue history.

The model output is
\begin{equation}
y_t=\{r_t,c_t,f_t\},
\label{eq:output}
\end{equation}
and is encoded with the fixed fields \texttt{\textless think\textgreater}$r_t$\texttt{\textless/think\textgreater}, \texttt{\textless answer\textgreater}$c_t$\texttt{\textless/answer\textgreater}, and \texttt{\textless function\textgreater}$f_t$\texttt{\textless/function\textgreater}. Here, $r_t$ describes observable visual evidence, the state assessment, and action prerequisites; $c_t$ is a concise operator-facing conclusion; and $f_t$ is the single skill to invoke. This explicit text is the model's chain-of-thought reasoning; exposing it to the operator lets a clinician inspect and verify why the particular skill was selected, which to some extent improves the interpretability of the decision. We caution that this rationale is a model-generated summary and is not verified to faithfully reflect the model's internal reasoning; we therefore present it as an auditable decision basis rather than as evidence of the model's actual decision process.

The model is restricted to the seven actions in Table~\ref{tab:skills}. Target localization, pose correction, and electrode pickup are deterministic internal stages of \skill{SPATIAL\_REGISTRATION}, \skill{ALIGN\_NEEDLE}, or \skill{INSERT}, rather than additional labels. Manual confirmation is a parser fallback and is not part of the model action set. Each skill is represented as
\begin{equation}
A_k=\langle n_k,P_k,C_k,T_k\rangle,
\end{equation}
where $n_k$ is its name, $P_k$ its prerequisite, $C_k$ its execution program, and $T_k$ its completion or termination criterion.

The model output first enters a safety parser. The parser verifies that each field appears exactly once, only one legal function is present, the function belongs to the seven-class set, and $P_k(s_t,I_t)$ holds. A malformed output, illegal or multiple function, or motion request in a high-risk state is never dispatched. The system observes again and requests a new inference, entering manual confirmation when necessary. Images and states are reacquired after every skill, forming an observe--reason--verify--execute--observe loop.

\begin{figure*}[t]
\centering
\includegraphics[width=\textwidth]{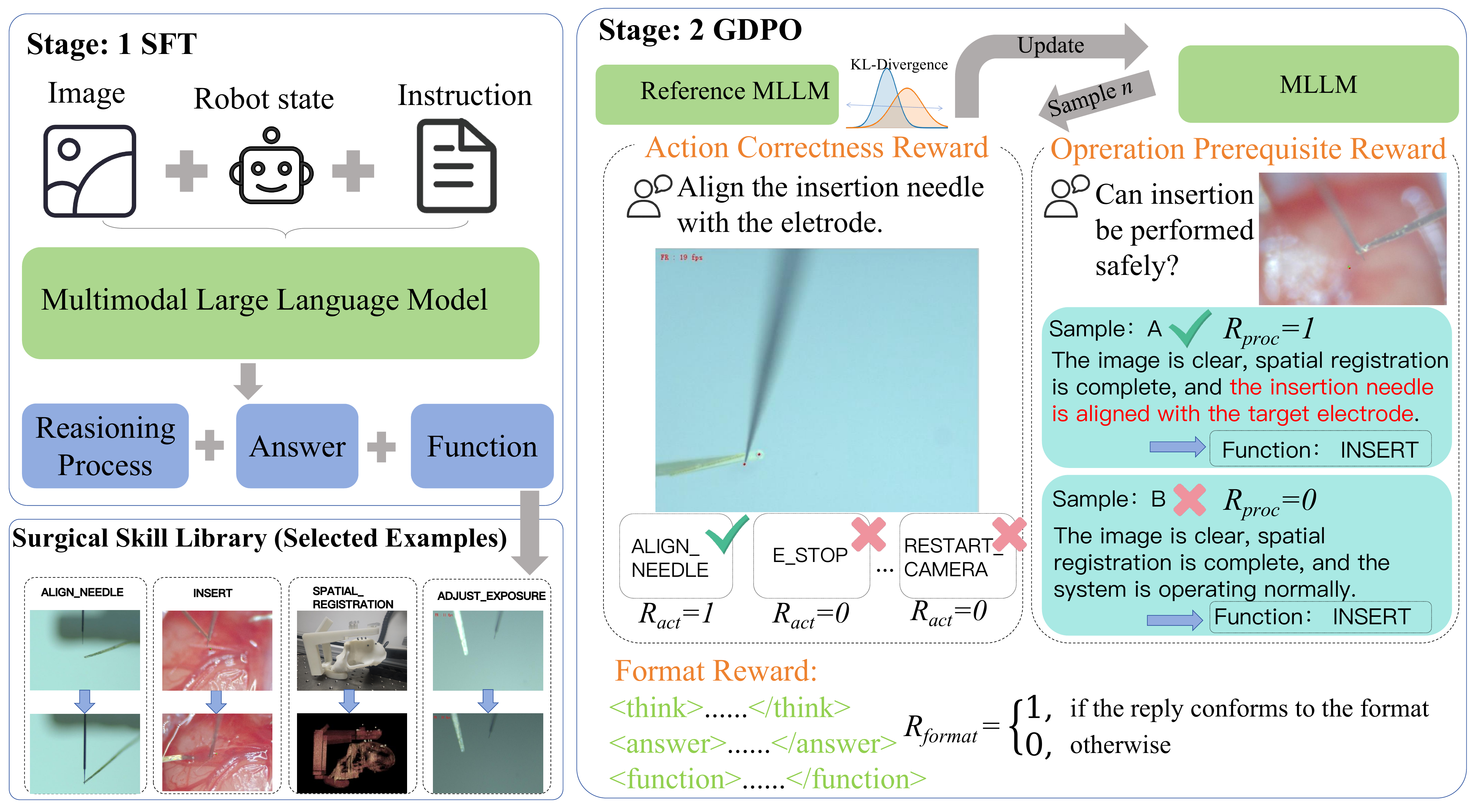}
\caption{SFT and GDPO training pipeline. The MLLM first learns structured reasoning, answers, and function calls through SFT. GDPO then optimizes action correctness, operation prerequisites, and output format under a reference-model constraint.}
\label{fig:training}
\vspace{-0.5em}
\end{figure*}

\subsection{Structured Data Generation and Partitioning}

Each sample is
\begin{equation}
d_i=\{I_i,s_i,q_i,r_i,c_i,f_i\}.
\end{equation}
Original samples comprise robot-platform images together with synchronized robot-state (joint-value) records, covering scenarios such as normal operation, image-quality anomalies, incomplete registration, camera failures, singular configurations, and needle misalignment. Visual perturbations include blur, exposure, noise, local occlusion, and target-region variation; state perturbations perturb the joint values themselves. Image and state perturbations are applied within consistent templates so that the visual evidence and the measured robot state do not contradict each other.

Before any augmentation, a grouped split is performed on the original sample IDs. With random seed 7, the 5,000 original samples are divided into 4,000 training sources, 500 validation sources, and 500 test sources. Each original sample and every augmented descendant generated from it are assigned exclusively to the same split. Sixfold augmentation is then applied independently in each split, so that every original sample yields six samples including the original (i.e., five augmented descendants), producing 24,000, 3,000, and 3,000 samples, respectively, for 30,000 in total. The validation and test sets use the same perturbation types as training, but no original sample or augmented descendant is shared across splits. Thus, the final 8:1:1 ratio applies to both original source groups and augmented samples. 

As shown in Fig.~\ref{fig:data}, to construct structured rationales, we use the same prompt template to query DeepSeek-VL2, Qwen3-VL-Flash, and ChatGPT-5.5 through API calls, with each model independently generating one structured rationale for each input. This multi-MLLM strategy leverages complementary model behaviors and reduces model-specific pattern bias in the generated dataset. All generators use the same seven-function vocabulary and output format, with temperature 0.7, top-$p$ 0.9, and a maximum output length of 512 tokens. Rule-based validation first checks tag completeness, function uniqueness, state consistency, and explicit safety rules.

Candidates that pass rule-based validation are cross-verified by the same three MLLMs in a leave-one-generator-out manner. Each model scores only the candidates generated by the other two models, rather than its own outputs; for example, Qwen3-VL-Flash scores only candidates generated by DeepSeek-VL2 and ChatGPT-5.5. This design reduces self-preference bias during verification. Each verifier assigns scores from 0 to 2 for visual consistency, state consistency, rationale quality, and function correctness, weighted by 0.30, 0.25, 0.20, and 0.25. A candidate is retained if the mean score is at least 1.6 and the verifier-score difference is no greater than 0.4. The final generated dataset is further audited through a random human spot check by two reviewers. A generated batch is considered valid only when the human acceptance rate exceeds 98\%; in our audit, 299 of 300 sampled instances were accepted (99.7\%).

\subsection{Supervised Fine-Tuning and GDPO}

\subsubsection{Supervised Fine-Tuning}
The base model is Qwen3-VL-4B-Instruct \cite{bai2025qwen3}. The autoregressive objective is
\begin{equation}
\mathcal{L}_{\mathrm{SFT}}=-\sum_{k=1}^{T}\log\pi_\theta(y_k\mid x,y_{<k}).
\end{equation}
LoRA\cite{hu2022lora} (rank 8, $\alpha=16$) is applied to all linear layers, with global batch 16, per-GPU micro-batch 1, maximum sequence length 8192, and learning rate $1\times10^{-5}$ (cosine schedule, 0.03 warmup, five epochs), using bfloat16, FSDP2\cite{fsdp}, gradient checkpointing, and eight NVIDIA A800-SXM4-80GB GPUs, with validation every 250 steps.

\subsubsection{Three GDPO Rewards}
For input $x$, the policy samples $G=4$ candidates,
$\{\hat y_j\}_{j=1}^{G}\sim\pi_\theta(\cdot\mid x)$.
Each candidate is scored by three binary rewards: action correctness,
operation prerequisites, and format validity. Because these rewards encode
distinct and partially competing objectives, they are combined with group
reward-decoupled normalization policy optimization (GDPO)\cite{liu2026gdpo},
which normalizes each reward independently within the sampled group before
aggregating them into the combined advantage.

The action-correctness reward evaluates whether the predicted function
matches the target atomic skill for the current image, robot state, and
operation query:
\begin{equation}
R_{\mathrm{act}}(\hat y_j\mid x)=
\mathbf{1}[\hat f_j\in\mathcal{F}^{+}(x)],
\end{equation}
where $\mathcal{F}^{+}(x)$ is the singleton set containing the single correct function call for the current input. For example, when needle alignment is the correct next step, only \skill{ALIGN\_NEEDLE} receives this reward.

The operation-prerequisite reward checks whether the current state satisfies the prerequisites of the predicted function; it therefore acts as a safety/precondition constraint rather than as a correctness signal:
\begin{equation}
R_{\mathrm{proc}}(\hat y_j\mid x)=
\begin{cases}
1,&P_{\hat f_j}(I,s)=1,\\
0,&P_{\hat f_j}(I,s)=0.
\end{cases}
\end{equation}
For example, \skill{INSERT} receives this reward only when the image is
usable, registration is complete, the camera is online, the robot is away
from singularity, and the needle is aligned. These prerequisites are
assessed by the system from the state $s_t$ (registration, camera,
singularity) and visual detection from $I_t$ (image usability, needle
alignment), all of which are available to the model as its input.

The format-validity reward is 1 when the output contains complete \texttt{think}, \texttt{answer}, and \texttt{function} fields with exactly one legal function call, and 0 otherwise.

Let $R_k^{(j)}\in\{0,1\}$ denote the value of reward
$k\in\{\mathrm{act},\mathrm{proc},\mathrm{fmt}\}$ for candidate $j$. GDPO
first normalizes each reward independently over the group, decoupling the
normalization of the three objectives as
\begin{equation}
\tilde A_k^{(j)}=
\frac{R_k^{(j)}-\mathrm{mean}\{R_k^{(1)},\dots,R_k^{(G)}\}}
{\mathrm{std}\{R_k^{(1)},\dots,R_k^{(G)}\}},
\label{eq:gdpo_group}
\end{equation}
where the mean and standard deviation are computed over the $G$ candidates
of the group and, when a reward is constant within the group (zero
variance), its normalized advantage is set to zero to avoid an undefined
division. The decoupled advantages are then combined with the reward
weights, which are applied to the normalized values rather than to the raw
reward scores:
\begin{equation}
\tilde A^{(j)}=
0.55\,\tilde A_{\mathrm{act}}^{(j)}+
0.35\,\tilde A_{\mathrm{proc}}^{(j)}+
0.10\,\tilde A_{\mathrm{fmt}}^{(j)}.
\label{eq:gdpo_weight}
\end{equation}
To keep the advantage magnitude stable as the number of rewards grows, we
apply a final batch-wise normalization over the minibatch $\mathcal{D}_B$:
\begin{equation}
\hat A^{(j)}=
\frac{\tilde A^{(j)}-\mathrm{mean}\{\tilde A^{(i)}\}_{i\in\mathcal{D}_B}}
{\mathrm{std}\{\tilde A^{(i)}\}_{i\in\mathcal{D}_B}+\epsilon}.
\label{eq:gdpo_batch}
\end{equation}
The policy is updated with a clipped surrogate objective on the normalized
advantages, with an importance ratio
$s_t(\theta)=\pi_\theta(y_t\mid x,y_{<t})/
\pi_{\theta_{\mathrm{old}}}(y_t\mid x,y_{<t})$ and a KL-toward-reference
regularizer $\beta\,\mathrm{KL}[\pi_\theta\|\pi_{\mathrm{ref}}]$:
\begin{equation}
\resizebox{\columnwidth}{!}{$\begin{aligned}
\mathcal{J}_{\mathrm{GDPO}}(\theta)
={}&\mathbb{E}\Big[\frac{1}{G}\sum_{j=1}^{G}\frac{1}{|y_j|}\sum_{t=1}^{|y_j|}\min\!\big(s_t\hat A^{(j)},\;\\
&\qquad\mathrm{clip}\!\big(s_t,1-\epsilon_c,1+\epsilon_c\big)\hat A^{(j)}\big)-\beta\,\mathrm{KL}\big[\pi_\theta\|\pi_{\mathrm{ref}}\big]\Big],
\end{aligned}$}
\label{eq:gdpo_obj}
\end{equation}
where $\pi_{\mathrm{ref}}$ is the SFT reference model, and $\epsilon_c=0.1$ and $\beta=0.01$ are the clipping threshold and KL coefficient, respectively. The
weights emphasize correct skill selection while retaining explicit penalty
for unsafe prerequisite violations and parser-invalid outputs. We adopt GDPO as defined in \cite{liu2026gdpo} without modification; the reward weights (0.55/0.35/0.10) are the only task-specific configuration.

\section{Experimental Design}
\subsection{Task and Data Split}
Experiments use closed-loop flexible-electrode implantation as the representative task. The model selects the next action from seven atomic skills; target localization, pose correction, and electrode pickup are performed internally by controllers. Before augmentation, the 5,000 original samples are grouped by original sample ID and split using seed 7 into 4,000/500/500 sources. All descendants of an original sample remain in the same split, after which independent within-split augmentation produces 24,000/3,000/3,000 samples. All offline models use the same fixed test set, which is excluded from training, hyperparameter selection, and reward design.

\subsection{Metrics and Statistical Protocol}
Offline skill selection is treated as seven-class classification. Accuracy is the proportion of samples with the correct function name. Precision, recall, and F1 are computed per class and macro-averaged so that frequent actions such as \skill{ADJUST\_EXPOSURE} do not obscure minority safety actions. Illegal, missing, or multiple functions are counted as incorrect predictions. Safety violation is evaluated over 2,529 test samples with explicit safety prerequisites and counts a motion command or \skill{INSERT} proposed when the image is unusable, registration is incomplete, the needle is misaligned, the camera is offline, self-check fails, or the robot approaches singularity. Parser-blocked violations are never sent to the robot.

Physical subtasks report success, positional error, orientation error, and time. Spatial registration uses target registration error (TRE), the post-registration distance between corresponding targets in image and robot coordinates. Alignment and insertion report orientation and positional errors separately; positional error is not added to TRE. Real-robot task success requires the system to execute all prescribed atomic skills and reach the terminal state. Mean stage count uses all trials. Offline metrics are reported as descriptive point estimates over the fixed augmented test set because samples derived from the same original ID are not independent observations. Binomial success rates in physical experiments use Wilson 95\% confidence intervals\cite{wilson1927}, and paired real-robot methods use two-sided McNemar tests\cite{mcnemar1947}.

\subsection{Robot Platform and Material Scope}
The real-robot platform follows the system in \cite{ye2026fully}: a UR5 arm performs large-range motion while the implantation head handles precise local motion, with two $2\times$ microscope cameras (MD65-2T26X lens, $2048\times2048$), a voice-coil-actuated insertion needle, a KOHZU XA04A-R102 stage (0.5-$\mu$m resolution, $\pm0.3~\mu$m repeatability), and a KOHZU SA07A-RM02 electrode stage ($0.0012^{\circ}$/$0.0015^{\circ}$ about $x$/$y$). A measured 2.4-mm field of view gives an image-plane scale of $2.4/2048=0.001171875$~mm/px, used only for local image residuals.

Success requires TRE $<0.50$~mm for spatial registration and positional error $<0.10$~mm for the controller-internal target-localization stage. \skill{ALIGN\_NEEDLE} requires orientation error $<2^{\circ}$ and image-plane error $<0.005$~mm; \skill{INSERT} requires orientation error $<5^{\circ}$ and three-dimensional landing error $<0.030$~mm. These thresholds evaluate the corresponding controller-level subtasks and are not used to define full-task completion. Spatial registration, target localization, image adjustment, and safety stopping use phantoms, calibration targets, and the robot platform without animal material. Only the physical \skill{INSERT} subtask and complete closed-loop implantation trials used legally sourced ex-vivo rat brain tissue obtained under institutional ethical approval and reported in accordance with the ARRIVE guidelines\cite{arrive2020}. 
\begin{table*}[t]
\caption{Atomic-Skill Selection on the Fixed Test Set}
\label{tab:offline}
\centering
\small
\resizebox{\textwidth}{!}{%
\begin{tabular}{lccccc}
\toprule
\textbf{Method} & \textbf{Accuracy (\%)\up} & \textbf{Macro-P (\%)\up} & \textbf{Macro-R (\%)\up} & \textbf{Macro-F1 (\%)\up} & \textbf{Safety violation (\%)\down} \\
\midrule
\multicolumn{6}{l}{\textit{General VLMs (non-SFT)}} \\
MiniCPM-o-2.6\cite{yao2024minicpm} & 14.2 & 2.0 & 14.3 & 3.6 & 18.6 \\
DeepSeek-VL2\cite{wu2024deepseek} & 20.6 & 22.4 & 20.9 & 14.8 & 14.2 \\
Llama-4-Maverick-17B-128E-Instruct\cite{llama4} & 24.7 & 28.9 & 25.1 & 17.2 & 11.6 \\
InternVL3.5-241B-A28B\cite{internvl35} & 31.6 & 37.8 & 31.2 & 24.4 & 7.4 \\
Qwen2.5-VL-3B-Instruct\cite{qwen25vl} & 15.4 & 27.8 & 15.5 & 5.9 & 15.8 \\
Qwen2.5-VL-7B-Instruct\cite{qwen25vl} & 28.0 & 29.7 & 28.2 & 18.2 & 10.9 \\
Qwen3-VL-4B-Instruct\cite{bai2025qwen3} & 30.0 & 40.8 & 30.3 & 23.8 & 7.8 \\
Qwen3-VL-30B-A3B-Instruct\cite{bai2025qwen3} & 35.0 & 44.1 & 35.2 & 27.6 & 5.9 \\
Qwen3-VL-235B-A22B-Thinking\cite{bai2025qwen3} & 38.4 & 43.6 & 38.0 & 31.2 & 4.8 \\
\midrule
\multicolumn{6}{l}{\textit{Surgical-specific VLMs}} \\
SurgVLM-7B\cite{zeng2025surgvlm} & 41.8 & 47.2 & 39.7 & 38.6 & 5.3 \\
SurgVidLM\cite{wang2025surgvidlm} & 44.6 & 49.1 & 43.8 & 42.7 & 4.9 \\
\midrule
\multicolumn{6}{l}{\textit{General VLMs + SFT}} \\
Qwen3-VL-4B-Instruct + SFT & 83.0 & 81.4 & 81.0 & 81.1 & 3.2 \\
Qwen3-VL-32B-Instruct + SFT & 89.1 & 87.6 & 85.6 & 86.5 & 1.9 \\
DeepSeek-VL2-Small + SFT & 81.7 & 78.7 & 78.0 & 78.0 & 4.2 \\
\midrule
\multicolumn{6}{l}{\textit{SFT + GDPO}} \\
\textbf{SurgiCLAW} & \textbf{94.6} & \textbf{94.0} & \textbf{92.0} & \textbf{93.0} & \textbf{0.5} \\
\bottomrule
\end{tabular}}
\vspace{-0.5em}
\end{table*}
\section{Results and Analysis}
\subsection{Offline Model Comparison}
All models are compared under one offline protocol. General VLMs without SFT estimate the zero-shot ceiling of generic multimodal ability; general VLMs with SFT quantify the benefit of state--skill labels; surgical VLMs test whether domain pretraining can replace task adaptation; and SurgiCLAW uses SFT followed by GDPO. Every model receives identical images, states, and queries and must use the same structured interface, which imposes a strict output format; prompt-only results therefore partly reflect adaptation to this interface rather than task ability alone.

On the fixed augmented test set, SurgiCLAW reaches 94.6\% accuracy and 93.0\% Macro-F1, improving over the strongest SFT general model by 5.5/6.5 points and over Qwen3-VL-4B-Instruct + SFT by 11.6/11.9 points, while its safety-violation rate drops to 0.5\% (13/2,529) versus 3.2\% and 1.9\% for the 4B and 32B SFT models. Prompt-only general VLMs reach only 14.2--38.4\% accuracy (Qwen3-VL-235B strongest at 31.2\% Macro-F1), indicating that scale alone does not solve the state-to-skill mapping. After SFT the 4B model rises to 83.0\%/81.1\%, the 32B to 89.1\%/86.5\%, and surgical-specific VLMs remain below SFT, showing that domain semantics do not replace interface adaptation. Because augmented samples derived from the same original ID are correlated, these offline results are descriptive point estimates without confidence intervals or significance tests.
\vspace{-0.5em}
\subsection{Real-Robot Comparison}

\begin{figure*}[t]
\centering
\includegraphics[width=\textwidth]{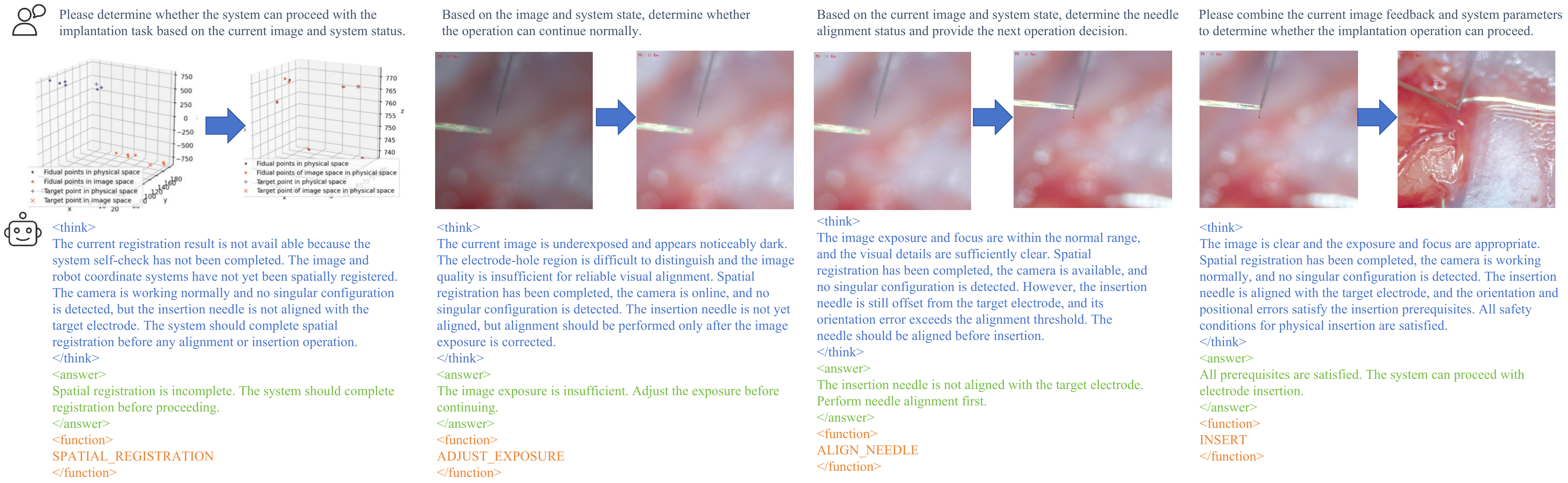}
\caption{Closed-loop decision example for flexible-electrode implantation. At each step, SurgiCLAW receives the current microscopic image, robot state, and operation query, then outputs a structured rationale, an operator-facing answer, and a single atomic-skill call. The example illustrates a typical decision sequence from spatial registration and image-quality correction to needle alignment and final insertion.}
\label{fig:demo}
\vspace{-1em}
\end{figure*}

Real-robot experiments evaluate the complete loop of high-level selection, deterministic control, and state update. SurgVLM-7B, SurgVidLM, Qwen3-VL-4B-Instruct + SFT, Qwen3-VL-32B-Instruct + SFT, and SurgiCLAW receive the same microscopic images, robot states, and task queries and share the same parser, action vocabulary, and low-level controllers. Each method is evaluated on 50 paired initial conditions, with randomized method order to reduce platform drift. Fig.~\ref{fig:demo} illustrates a representative closed-loop decision sequence, showing how SurgiCLAW maps successive image and robot-state observations to atomic-skill calls.

\begin{table*}[t]
\caption{Closed-Loop Real-Robot Implantation Results}
\label{tab:robot}
\centering
\scriptsize
\setlength{\tabcolsep}{3.5pt}
\renewcommand{\arraystretch}{0.92}
\resizebox{0.80\textwidth}{!}{%
\begin{tabular}{lcccc}
\toprule
\textbf{Method} & \textbf{Success (\%)\up} & \textbf{Stages\down} & \textbf{Takeover (\%)\down} & \textbf{Unsafe proposal (\%)\down} \\
\midrule
\multicolumn{5}{l}{\textit{Surgical-specific VLMs}} \\
SurgVLM-7B & 46.0 (23/50) & $10.4\pm2.6$ & 38.0 (19/50) & 5.4 (11/204) \\
SurgVidLM & 50.0 (25/50) & $9.8\pm2.4$ & 34.0 (17/50) & 4.9 (10/205) \\
\midrule
\multicolumn{5}{l}{\textit{General VLM + SFT}} \\
Qwen3-VL-4B-Instruct + SFT & 66.0 (33/50) & $8.6\pm1.9$ & 20.0 (10/50) & 3.4 (7/207) \\
Qwen3-VL-32B-Instruct + SFT & 76.0 (38/50) & $7.7\pm1.6$ & 10.0 (5/50) & 1.9 (4/206) \\
\midrule
\multicolumn{5}{l}{\textit{SFT + GDPO}} \\
\textbf{SurgiCLAW} & \textbf{88.0 (44/50)} & $\mathbf{6.9\pm1.5}$ & \textbf{4.0 (2/50)} & \textbf{0.5 (1/194)} \\
\bottomrule
\end{tabular}}
\vspace{-1em}
\end{table*}

\begin{table*}[t]
\caption{Ablation of Model-Side Components}
\label{tab:ablation}
\centering
\small
\begin{tabular}{lccccc}
\toprule
\textbf{Setting} & \textbf{Accuracy (\%)\up} & \textbf{Macro-P (\%)\up} & \textbf{Macro-R (\%)\up} & \textbf{Macro-F1 (\%)\up} & \textbf{Safety violation (\%)\down} \\
\midrule
\textbf{Full model} & \textbf{94.6} & \textbf{94.0} & \textbf{92.0} & \textbf{93.0} & \textbf{0.5} \\
w/o structured rationale & 88.2 & 87.4 & 85.9 & 86.4 & 2.3 \\
w/o GDPO & 83.0 & 81.4 & 81.0 & 81.1 & 3.2 \\
w/o action-correctness reward & 90.4 & 88.1 & 88.7 & 88.2 & 1.6 \\
w/o prerequisite reward & 91.2 & 90.8 & 88.9 & 89.7 & 3.2 \\
w/o format reward & 92.0 & 92.8 & 88.6 & 90.4 & 1.1 \\
\bottomrule
\end{tabular}
\vspace{-0.5em}
\end{table*}
Table~\ref{tab:robot} reports closed-loop results. The safety-violation denominator is the number of safety-critical decision opportunities across trials, which differs by method because failed or repeated loops expose different numbers of such states; every violation is blocked by the parser. SurgiCLAW completes 44/50 trials (Wilson 95\% interval 76.2--94.4\%), improving over Qwen3-VL-4B-Instruct + SFT by 22 points (33/50 to 44/50) with 19.8\% fewer stages and takeover falling from 20.0\% to 4.0\%; a paired McNemar test over this pair is statistically significant ($p<0.05$). Relative to Qwen3-VL-32B-Instruct + SFT, it improves by 12 points (38/50 to 44/50), but this difference is not statistically significant (two-sided McNemar $p=0.070$), so we treat the gain over the 32B model as a trend rather than a confirmed improvement at 50 trials. Because all methods share the same low-level controllers, differences arise from the high-level policy triggering registration, alignment, insertion, recovery, and stopping at the right state: the surgical VLMs delay registration, repeat alignment, or request insertion before prerequisites hold, while the 32B SFT model is more robust than the 4B model, which is more sensitive to ambiguous visual states and recovery decisions. Failure categories are high-level selection, repeated alignment without convergence, visual anomaly/occlusion, safety-parser refusal, and insertion-control failure; total failures are 27/25/17/12/6, and SurgiCLAW's six failures break down as 1/2/1/1/1, so repeated alignment under occlusion and physical insertion remain the principal bottlenecks.
\vspace{-0.5em}
\subsection{Ablation Study}
Ablations remove only model-side training or output components, while keeping low-level controllers, data splits, and metrics unchanged. All variants are evaluated on the same fixed test set with accuracy, macro-precision, macro-recall, Macro-F1, and safety violation.

Table~\ref{tab:ablation} shows the ablation results. Removing the structured rationale lowers accuracy/Macro-F1 by 6.4/6.6 points, and the SFT-only variant (w/o GDPO, equal to Qwen3-VL-4B-Instruct + SFT) remains far above prompt-only baselines but below the full model, indicating that supervised skill labels alone do not fully cover boundary and safety-critical states: adding GDPO raises accuracy from 83.0\% to 94.6\%, Macro-F1 from 81.1\% to 93.0\%, and cuts safety violation from 3.2\% to 0.5\%. Removing the action-correctness reward mainly degrades class-level selection, removing the prerequisite reward gives the highest safety-violation rate (confirming its role in constraining state/function conflicts), and removing the format reward mainly lowers recall. GDPO therefore improves accuracy, parseability, and safety-aware selection.

\subsection{Controller-Level Subtask Verification}
Each deterministic controller was tested 40 times under fixed perturbations, independently of the large-model decisions, following the frameworks of \cite{ye2024non,ye2026fully,ye2026holenet}.

\begin{table*}[t]
\caption{Atomic-Subtask Success and Execution Accuracy}
\label{tab:subtasks}
\centering
\small
\resizebox{\textwidth}{!}{%
\begin{tabular}{lllll}
\toprule
\textbf{Subtask} & \textbf{Success criterion} & \textbf{Success (\%)\up} & \textbf{Mean error\down} & \textbf{Time (s)\down} \\
\midrule
Spatial registration & TRE $<0.50$ mm and coordinate transform completed & 95.0 (38/40) & $0.347\pm0.126$ mm (TRE) & $3.18\pm0.37$ \\
Target localization (internal) & Stable target detection available for alignment & 97.5 (39/40) & $0.051\pm0.028$ mm (position) & $0.82\pm0.26$ \\
Needle alignment & Orientation and image-plane position within thresholds & 95.0 (38/40) & $0.91\pm0.42^{\circ}$ / $0.0018\pm0.0010$ mm & $19.4\pm4.8$ \\
Insertion execution & Direction and 3-D landing position within thresholds & 90.0 (36/40) & $2.57\pm1.16^{\circ}$ / $0.015\pm0.006$ mm & $45.73\pm13.49$ \\
Image-quality adjustment & Image restored to a parseable quality range & 97.5 (39/40) & -- & $1.63\pm0.61$ \\
Safety stop & Timely refusal in a high-risk state & 100.0 (40/40) & -- & $0.12\pm0.02$ \\
\bottomrule
\end{tabular}}
\vspace{-0.5em}
\end{table*}

Table~\ref{tab:subtasks} reports controller-level success and accuracy. All controllers
succeed on 90.0\%--100.0\% of trials; the spatial-registration TRE of $0.347\pm0.126$~mm,
needle-alignment error of $0.91\pm0.42^{\circ}$, and insertion error of $0.015\pm0.006$~mm
fall within the corresponding thresholds. Errors and times are means and standard deviations
over successful trials.

\section{Conclusion}
This paper presented SurgiCLAW, a hierarchical atomic-skill interface for precision robotic micromanipulation. Its central idea is to keep the semantic decision making of a multimodal language model within a small, bounded, and auditable skill space, and to gate every decision through a safety parser before delegating execution to verified deterministic controllers; this separates \emph{what} to do next from \emph{how precisely} to do it. It also makes the interface learnable from limited surgical data: a multi-MLLM data-generation pipeline expands 5,000 state--skill samples into 30,000 structured ones, and a three-reward group reward-decoupled normalization policy optimization objective improves skill selection, prerequisite safety, and output parseability. On the fixed test set SurgiCLAW reaches 94.6\% accuracy and 93.0\% Macro-F1, and it completes 88.0\% of 50 closed-loop implantation trials. We interpret these results as a feasibility demonstration rather than evidence of clinical efficacy: the evaluation covers a single high-precision task using phantom and ex-vivo material, and the closed-loop comparisons are constrained by the modest number of paired trials. Whether the safety and interpretability benefits generalize to other tasks, to more stringent clinical conditions, or to long-horizon unattended operation remains to be established.

\bibliographystyle{IEEEtran}
\bibliography{references}

\end{document}